\documentclass[11pt]{article}

\usepackage[preprint]{acl}

\usepackage{times}
\usepackage{latexsym}
\usepackage{amsmath}
\usepackage{amssymb}
\usepackage{booktabs}
\usepackage{subcaption} 

\usepackage[T1]{fontenc}

\usepackage[utf8]{inputenc}
\usepackage{kotex}

\usepackage{microtype}

\usepackage{inconsolata}

\usepackage{graphicx}
\usepackage{stfloats}
\usepackage{tcolorbox}
\usepackage{pgfplots}
\pgfplotsset{compat=1.18}

\title{Predicate Importance Estimation and Decoupled Rationale-Score Distillation for Entity Alignment}

\author{
 \textbf{Keunha Kim\textsuperscript{1}},
 \textbf{Yoonjin Jang\textsuperscript{1}},
 \textbf{Hyeon-gu Lee\textsuperscript{2}},
 \textbf{Sihyung Kim\textsuperscript{2}},
 \textbf{Youngjoong Ko\textsuperscript{1}}\thanks{Corresponding author}
\\
 \textsuperscript{1}SungKyunKwan University,
 \textsuperscript{2}NAVER,
\\
 \texttt{\{keunhakim98, yoonjinjang98\}@gmail.com, yjko@skku.edu}
\\
 \texttt{\{hyeongu.lee, sihyung.kim\}@navercorp.com}
}

\begin{document}
\maketitle 
\begin{abstract}
Knowledge graphs (KGs) are increasingly used as structured context for Large Language Models (LLMs), but industrial KG-RAG systems often need to integrate public and domain-specific KGs constructed from heterogeneous databases. This integration relies on Entity Alignment (EA), where lexical matching alone is insufficient under predicate-name variation and incomplete local neighborhoods. We address EA for KG integration by constructing a pairwise EA dataset and proposing two complementary modules: Predicate Importance Estimation (PIE) and Decoupled Rationale-Score Distillation (DRSD). PIE is a compact embedding-based approach that removes the subject information from each 1-hop triple, encodes the resulting subjectless triples, and aggregates them with learnable predicate-importance weights to build predicate-aware entity embeddings. DRSD trains a distilled small language model (SLM) with pseudo-answers produced by a teacher LLM through distinct prompts. By converting binary EA labels into text-based supervision and decoupling confidence-score estimation from label-consistent rationales, DRSD enables the SLM to learn task-specific reasoning while retaining a less label-biased confidence signal. Experiments show that PIE and DRSD improve EA classification. Moreover,
because DRSD decouples confidence-score estimation from the decision,
a discrepancy between the two flags an uncertain prediction for human review, thereby enabling a practical discrepancy between automatic acceptance and human-in-the-loop verification.
\end{abstract}

\section{Introduction}

Knowledge graphs (KGs) are increasingly used as structured knowledge sources for Large Language Models (LLMs) \cite{pan2024unifying,sun2024thinkongraph}. Industrial systems simultaneously reason over public KGs such as DBpedia and Wikidata, as well as domain-specific KGs that are constructed from multiple internal relational databases via standard mappings \cite{w3c2012r2rml,sequeda2019pay}. Therefore, these heterogeneous sources need to be integrated into a unified KG.
To build such a unified KG, duplicate entities must be identified and merged across sources, which requires an explicit entity-alignment step. \textbf{Entity Alignment} (EA) identifies equivalent entities across distinct KGs
\cite{sun2020openea,zhang2022easurvey}. EA is intrinsically hard for two
reasons that matter especially when systems bridge different
graphs. First, two graphs typically use different predicate-naming
conventions (e.g., \texttt{employer\_of} vs.\ \texttt{worksFor}), so
purely structural or string-based matching is misled by surface
variance. Second,
knowledge graphs collected for different purposes often contain different types and amounts of information. 
Therefore, robust EA must go beyond semantic
similarity through joint modeling of the surrounding subgraph and the
\textbf{relative importance} of each predicate as evidence for alignment.

\begin{figure}[t]
  \centering
  \includegraphics[width=\columnwidth]{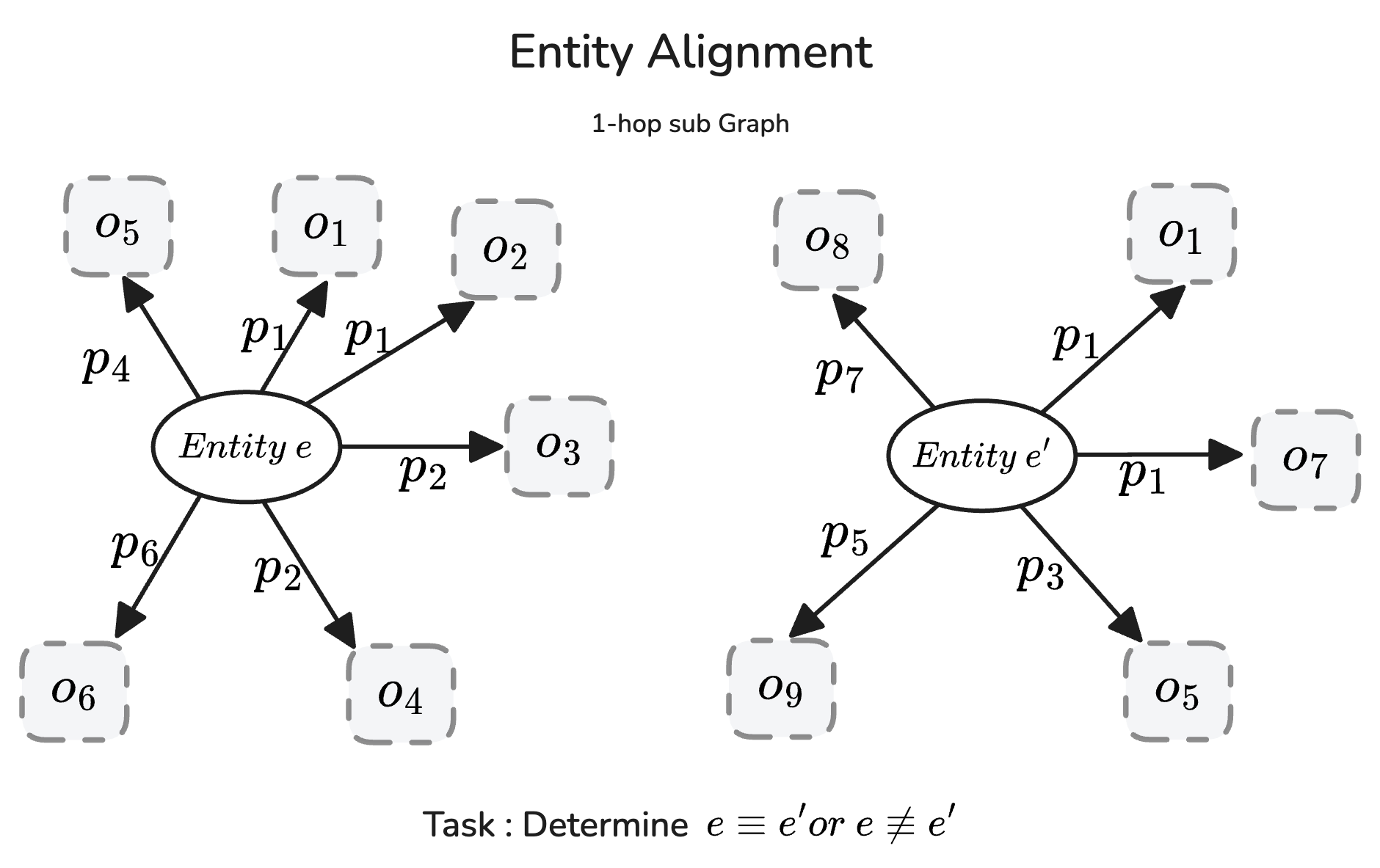}
  \caption{1-hop graph example for an entity-alignment pair}
  \vspace{-0.2em}
  \label{fig:task}
\end{figure}

To handle such complex semantic judgments, Large Language Models (LLMs) offer attractive generalization capabilities, and these capabilities have led recent work to exploit them for EA \cite{jiang2024chatea,zhang2024autoalign,chen2024llm4ea}. However, these LLM-based aligners rely heavily on general pretraining knowledge. As a result, they struggle to internalize the very domain-specific relations and user-defined alignment constraints that define these heterogeneous graphs because such domain-specific patterns were under-represented during pretraining.

In this work, we address entity alignment for KG integration by constructing a pairwise EA dataset and introducing two complementary alignment modules. \textbf{Our dataset} reflects the pre-alignment setting: each instance contains two ambiguous candidate entities and their 1-hop subgraph evidence, so the model must decide equivalence using local graph context.

\textbf{Our first module,} \textbf{Predicate Importance Estimation} (PIE), is a compact embedding-based approach that removes the subject information from each 1-hop triple, independently encodes the subjectless triples, and aggregates the representations with learnable predicate-importance weights. As a result, PIE constructs a compact subgraph-level entity embedding, estimates which predicates are most useful for EA, and achieves strong entity alignment performance.

\textbf{Our second module,} \textbf{Decoupled Rationale-Score Distillation (DRSD)}, trains a distilled small language model (SLM) using decoupled rationale-score pseudo-answers produced by an LLM from distinct prompts. Since binary labels lack the reasoning and confidence signals needed for text-generating models, DRSD uses task-specific prompts to elicit structured teacher responses and convert EA labels into text-based supervision. When eliciting responses from the teacher LLM through prompts, the gold-label prompt produces a label-consistent decision and rationale, whereas the score prompt asks the LLM to estimate similarity without exposing the gold label or requiring a generated decision. The final SFT target combines the gold-label decision and justification with the score prompt confidence score, allowing the student model to learn task-specific reasoning while retaining a less label-biased confidence signal.

Compared with the prompt-only LLM baseline, both proposed modules improve EA performance when trained on the constructed dataset. PIE improves Acc. and F1 by 6.32 and 4.63 percentage points, and DRSD improves them by 7.66 and 6.76 percentage points. Moreover, because DRSD decouples the confidence score from the decision, disagreements between the two can flag predictions to route to human review (13.3\% of pairs), which can further increase F1 from 0.886 to 0.909.
 \section{Related Work}

\subsection{Entity Alignment and LLM-based Approaches}

Entity Alignment (EA) has been studied through embedding-based and
neighborhood-aware methods, including translation-style KG embeddings,
cross-graph extensions, and GNN-based aligners
\cite{sun2020openea,zhang2022easurvey,chen2017mtranse,sun2018bootea,wang2018gcnalign,ijcai2019p733,sun2020alinet}.
These methods show that structural, attribute, and predicate-aware
features are important for distinguishing equivalent entities across
heterogeneous KGs \cite{trisedya2019attrea,xu2019subgraph,zhang2022easurvey}.
More recent LLM-based approaches, such as ChatEA, AutoAlign, and
LLM4EA, exploit the semantic reasoning ability of LLMs for EA
\cite{jiang2024chatea,zhang2024autoalign,chen2024llm4ea}. However,
they can be brittle on domain-specific relations and costly to run for
each candidate pair \cite{ling2023domainllm}. Our work complements
these lines of research by learning predicate-aware entity
representations and distilling LLM alignment reasoning into a smaller
model with a score-based confidence signal.

\subsection{Knowledge Distillation from LLMs to Small Models}

Knowledge distillation \cite{hinton2015distillation} has recently
been extended from logit-matching to rationale transfer.
\textbf{Distilling Step-by-Step} \cite{hsieh2023distilling} trains a
small model jointly on LLM-generated labels and chain-of-thought
rationales \cite{NEURIPS2022_9d560961} to match teacher performance with far
fewer parameters. Subsequent work
\cite{magister2023teaching,ho2023cotdistill} demonstrates that small
models can be taught to reason via CoT distillation, and SCOTT
\cite{wang2023scott} introduces a counterfactual objective to reduce shortcut imitation.

\subsection{Predicate-Aware KG Embeddings}

Beyond the translation-based and GNN-based methods above, several
lines of work explicitly model relation or attribute importance.
KBGAT \cite{nathani2019kbat} introduces attention-based neighborhood weighting for embeddings; relation-aware attention is similarly central to MRAEA \cite{mao2020mraea} and RDGCN \cite{ijcai2019p733};
and attribute-centric methods such as JAPE \cite{sun2017jape} and
AttrE \cite{trisedya2019attrea} treat predicates and literal
attributes as first-class signals.
PIE instead uses subjectless, predicate-aware representations
with importance-weighted pooling to produce a compact entity signature.

\section{Preliminaries}

Let $\mathcal{C}=\{G_1,\dots,G_M\}$ denote a collection of knowledge graphs (KGs). Each KG is defined as
\begin{equation}
G_m=(\mathcal{E}_m,\mathcal{P}_m,\mathcal{T}_m),
\end{equation}
where $\mathcal{E}_m$ is the entity set, $\mathcal{P}_m$ is the predicate set, and $\mathcal{T}_m$ is the triple set of the $m$-th KG. We assume that entities are unique within each KG, while equivalent entities may appear across different KGs.

Our task is to compare entities from two different KGs, $G_m$ and $G_n$ ($m\neq n$), and decide whether they denote the same entity. E.g., $e\equiv e'$ or $e\not\equiv e'$, $e \in G_m, e' \in G_n$. Each triple is denoted as $(s,p,o)$, where $s$, $p$, and $o$ denote subject, predicate, and object, respectively. For an entity $e\in\mathcal{E}_m$, its outgoing 1-hop neighborhood $\mathcal{N}_m(e)$ is
\begin{equation}
\mathcal{N}_m(e)=\{(s,p,o)\in\mathcal{T}_m \mid s=e\}.
\end{equation}
For a cross-KG entity pair $(e,e')\in\mathcal{E}_m\times\mathcal{E}_n$, we compute its similarity score $s_{e,e'}$ as
\begin{equation}
s_{e,e'}=\phi\big(\mathcal{N}_m(e),\mathcal{N}_n(e')\bigr),
\end{equation}
where $\phi(\cdot,\cdot)$ is an entity-pair scoring function. The final alignment decision $\widehat{y}_{e,e'}$ is then
\begin{equation}
\widehat{y}_{e,e'}=
\begin{cases}
1, & \text{if } s_{e,e'}\ge\tau, \\
0, & \text{otherwise},
\end{cases}
\end{equation}
where $\tau$ is a decision threshold, $\widehat{y}_{e,e'}=1$ indicates $e\equiv e'$, and $\widehat{y}_{e,e'}=0$ indicates $e\not\equiv e'$.

Thus the task is to retrieve similar cross-collection candidates and decide match/non-match for integration.

\section{Methodology}

Two complementary components are used for entity alignment. The first one is Predicate Importance Estimation (PIE), a lightweight statistical module that learns which predicates are most informative for entity alignment. The second one is Decoupled Rationale-Score Distillation (DRSD), a reasoning-aware distillation framework that transfers decision patterns from a large teacher model to a smaller student model.

\begin{figure*}[t]
  \centering
  \includegraphics[width=1\linewidth]{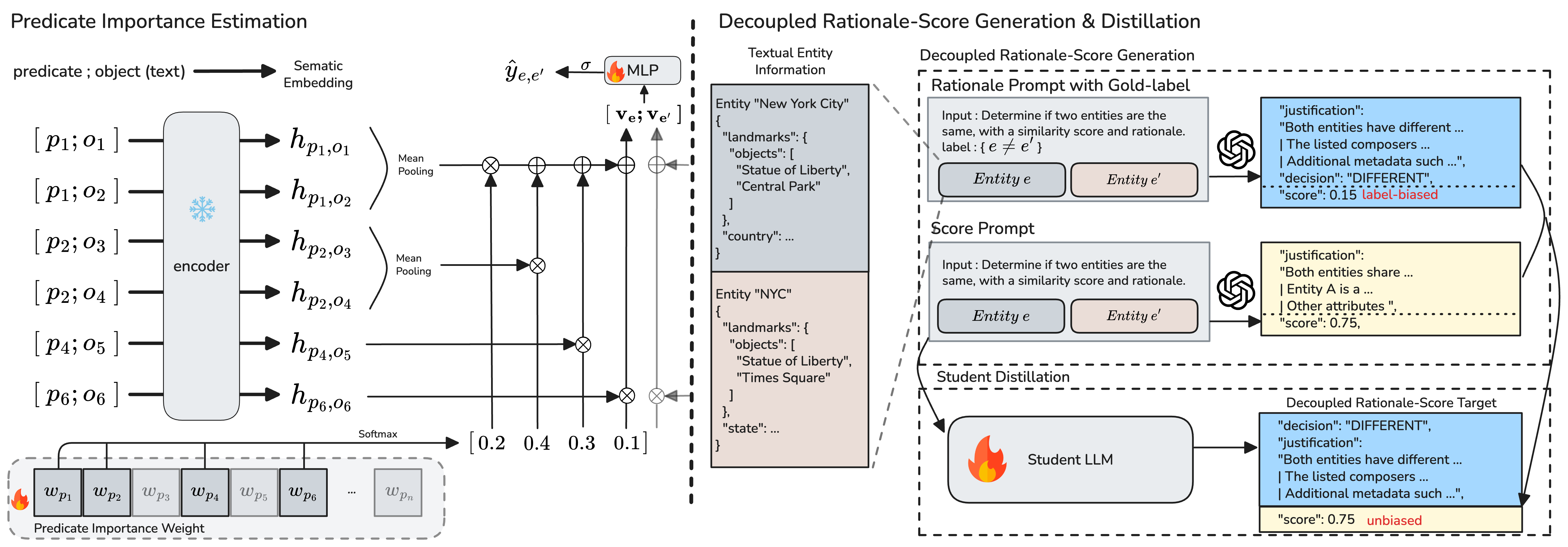}
  \caption{Overview of the proposed framework.  (Left) Entity embedding construction ($v_e$) within PIE; embeddings for each entity pair are subsequently concatenated for MLP-based classification. (Right) Illustrative example of the decoupled rationale-score distillation process.}
  \label{fig:framework_overview}
\end{figure*}

\subsection{Predicate Importance Estimation}

For the 1-hop neighborhood $\mathcal{N}_m(e)$ of an entity $e$, Predicate Importance Estimation (PIE) constructs a representative embedding $\mathbf{v}_e$ from local relational evidence. It first encodes subjectless triples, then pools triple representations by predicate, and finally aggregates predicate-level representations with predicate-importance weights. This process yields both an entity embedding and predicate-importance weights, and the weights indicate which predicates are most informative for alignment. PIE consists of three stages: (1) subjectless triple encoding, (2) predicate-wise pooling, and (3) predicate-importance pooling.

\paragraph{Subjectless triple encoding.} For each outgoing triple $(e,p,o)\in\mathcal{N}_m(e)$, we remove the subject and encode only the predicate--object text. This reduces reliance on the surface name of $e$ and makes entity comparison depend more on relational evidence:
\begin{equation}
\mathbf{h}_{p,o}^{(e)}=\mathrm{Enc}\big(\mathrm{text}(p)\oplus \mathrm{text}(o)\bigr),
\end{equation}
where $\oplus$ denotes text concatenation. Let $\mathcal{P}(e)=\{p\mid \exists o:(e,p,o)\in\mathcal{N}_m(e)\}$ be the set of predicates around $e$, and let $\mathcal{O}_p(e)=\{o\mid(e,p,o)\in\mathcal{N}_m(e)\}$ be the object set for predicate $p$.

\paragraph{Predicate-wise pooling.} Since an entity may have multiple triples with the same predicate, we apply mean pooling over their triple embeddings to obtain one predicate-level representation:
\begin{equation}
\mathbf{r}_{p}^{(e)}=\frac{1}{|\mathcal{O}_p(e)|}\sum_{o\in\mathcal{O}_p(e)}\mathbf{h}_{p,o}^{(e)}.
\end{equation}

\paragraph{Predicate-importance pooling.} We then aggregate the predicate-level representations into a single entity embedding:
\begin{equation}
\begin{aligned}
\alpha_p &= \frac{\exp(w_p)}{\sum_{p'\in\mathcal{P}(e)}\exp(w_{p'})}, \\
\mathbf{v}_e &= \sum_{p\in\mathcal{P}(e)} \alpha_p\mathbf{r}_{p}^{(e)}.
\end{aligned}
\end{equation}
Here, $w_p$ is a learnable scalar parameter for predicate $p$, and its softmax-normalized value $\alpha_p$ represents the relative importance of that predicate within $\mathcal{P}(e)$.
\paragraph{Entity-pair training.} For a candidate pair $(e,e')$, we concatenate the two entity embeddings and feed them into an MLP classifier:
\begin{equation}
\hat{y}_{e,e'}=\sigma\big(\mathrm{MLP}([\mathbf{v}_e;\mathbf{v}_{e'}])\bigr),
\end{equation}
where $\sigma(\cdot)$ is the sigmoid function and $\hat{y}_{e,e'} \in [0,1]$ denotes the predicted match probability. We train the classifier with binary cross-entropy over entity pairs with labels, where $y_{e,e'}=1$ indicates that $e$ and $e'$ are equivalent, and $y_{e,e'}=0$ otherwise. During training, we freeze the pretrained encoder and update only the predicate-importance weights and the MLP classifier layers. This keeps the training overhead low and quickly adapts the predicate-importance weights to the alignment task.

\subsection{Decoupled Rationale-Score Distillation}

Conventional LLM distillation trains students to imitate teacher outputs,
which makes the teacher an implicit performance ceiling when its responses
are incorrect. DRSD instead separates pseudo-answer generation into a
supervised signal, where the teacher sees the gold label, and an
unsupervised signal, where it does not. Combining the two preserves
label-guided reasoning while yielding a confidence score that is less
tied to the supervised answer.

\paragraph{Decoupled Rationale-Score Label Construction}
Our goal is to construct pseudo-answers that preserve the teacher's reasoning ability while preventing the confidence score from being biased toward the gold label in the prompt, regardless of the objective similarity between the two entities. For the $i$-th labeled entity pair, let $X_i$ denote the serialized evidence for the two entities and $y_i$ its gold label. We use three prompt schemas:
\begin{equation}
\begin{aligned}
q_i^{\mathrm{gold}} &= \mathrm{Prompt}_{\mathrm{gold}}(X_i,y_i), \\
q_i^{\mathrm{score}} &= \mathrm{Prompt}_{\mathrm{score}}(X_i), \\
q_i^{\mathrm{sft}} &= \mathrm{Prompt}_{\mathrm{sft}}(X_i).
\end{aligned}
\end{equation}
The gold-label prompt generates a supervised decision and rationale, whereas the score prompt estimates confidence from the evidence without receiving the gold label or producing a decision. The sft prompt is a simplified student training prompt for efficient fine-tuning. The teacher outputs are
\begin{equation}
\begin{aligned}
\mathbf{o}_i^{\mathrm{gold}} &= T(q_i^{\mathrm{gold}})
= \bigl(j_i^{\mathrm{gold}},\;d_i^{\mathrm{gold}},\;c_i^{\mathrm{gold}}\bigr), \\
\mathbf{o}_i^{\mathrm{score}} &= T(q_i^{\mathrm{score}})
= \bigl(j_i^{\mathrm{score}},\;c_i^{\mathrm{score}}\bigr),
\end{aligned}
\end{equation}
where $d$, $c$, and $j$ denote the decision, score, and justification fields, respectively. Appendix~\ref{sec:prompt_examples} provides the prompt templates and representative JSON-style teacher responses.

We then compose the final SFT target with the decision and justification from the gold-label output and the score from the score prompt output:
\begin{equation}
\ell_i^{\mathrm{sft}}=\bigl(d_i^{\mathrm{gold}},\; c_i^{\mathrm{score}},\; j_i^{\mathrm{gold}}\bigr).
\end{equation}
These three fields correspond to the final alignment decision, confidence score, and reasoning justification, respectively. This decoupled rationale-score construction uses label information where it is most reliable---the decision and justification---and keeps confidence estimation independent of the gold label in the prompt. As a result, the score can be used as a more objective thresholding signal for precision--recall trade-offs and human-in-the-loop review.

\subsubsection{Predicate-Importance Weights}
\label{sec:PIW}
The predicate-importance weights learned by PIE are also used to
compress LLM inputs. For a budget $k$, we keep only the triples whose
predicates fall in the top-$k$ set,
\begin{equation}
\mathrm{TopK}(\alpha, k) = \operatorname*{arg\,max}_{S \subseteq P(e),\, |S|=k} \sum_{p \in S} \alpha_p ,
\end{equation}
\begin{equation}
\tilde{\mathcal{N}}_m(e) = \{(e, p, o) \in \mathcal{N}_m(e) \mid p \in \mathrm{TopK}(\alpha, k)\} ,
\end{equation}
where $\alpha_p$ is the predicate-importance weight from Eq.~(7).
We use $k{=}10$ and further retain at most 10 objects per predicate
(Appendix~\ref{sec:PA construction}), which preserves informative evidence
while reducing prompt length and inference cost.

\subsubsection{SFT}

We construct an SFT dataset for a smaller language model as prompt--pseudo-answer pairs:
\begin{equation}
\mathcal{D}_{\mathrm{SFT}}
=\{(q_i^{\mathrm{sft}},\ell_i^{\mathrm{sft}})\}_{i=1}^{N}.
\end{equation}
The student model is fine-tuned to generate the decoupled rationale-score target response from the sft prompt. In this way, the small language model learns the gold-label decision and justification, and it inherits the prompt-derived score as its confidence signal. At deployment time, the generated score can be thresholded to trade precision against recall according to application constraints. To analyze the effect of score decoupling, we compare DRSD against \textbf{Label-Guided Distillation (LGD)}, which is an ablation that uses a single gold-label prompt to jointly produce the decision, rationale, and score without decoupling.

\section{Experiments}

\subsection{Datasets \& Models}

\paragraph{Datasets} We evaluate our method on a real-world, industrial-scale labeled entity-pair dataset constructed from our in-house heterogeneous knowledge graphs. The dataset contains a total of 85,858 training entity pairs and 5,000 evaluation pairs. Each pair is assigned a binary label: \textsc{same} for equivalent entities and \textsc{different} for non-equivalent entities. A detailed description of the underlying knowledge graphs and dataset statistics, including 1-hop triple distributions, is provided in Appendix~\ref{sec:data_statistics}.

\paragraph{Models}
For PIE, we use \texttt{BAAI/bge-m3}\footnote{\url{https://huggingface.co/BAAI/bge-m3}} to generate embeddings and employ a two-layer MLP as the classifier. For DRSD, we use \texttt{openai/gpt-oss-120b}\footnote{\url{https://huggingface.co/openai/gpt-oss-120b}} as the teacher LLM and train a \texttt{hcx-srch1-8B} (SLM) for distillation. Detailed architecture and training hyperparameters are provided in Appendix~\ref{sec:implementation_details}. 

\begin{table}[t]
\centering
\small
\setlength{\tabcolsep}{3pt}
\begin{tabular*}{\columnwidth}{@{\extracolsep{\fill}}lcccc@{}}
\toprule
\textbf{Method}
& \textbf{Acc.} & \textbf{Prec.} & \textbf{Rec.} & \textbf{F1}
\\
\midrule
LLM    & 0.7976 & 0.7433 & 0.9092 & 0.8179 \\
SLM    & 0.7418 & 0.7737 & 0.6836 & 0.7258 \\
\cmidrule(lr){1-5}
PIE(ours)    & 0.8608 & 0.8438 & 0.8856 & 0.8642 \\
DRSD(ours)  & 0.8742 & 0.8126 & 0.9728 & 0.8855 \\
\quad+ Human Review & \textbf{0.9020} & \textbf{0.8463} & \textbf{0.9824} & \textbf{0.9093} \\
\bottomrule
\end{tabular*}
\caption{Entity alignment performance on the evaluation split. LLM and
SLM are prompt-only; DRSD is fine-tuned from the SLM. \textit{+ Human
Review} corrects the 13.3\% of pairs flagged as score--decision
conflicts.}
\label{tab:main_result}
\end{table}

\subsection{Results}

Table~\ref{tab:main_result} reports entity alignment performance on the
evaluation split. For DRSD, the model outputs both a decision and a
similarity score. we compute the reported metrics from the decision
output by treating \textsc{same} predictions as the positive label.

The results highlight two main findings. First, PIE achieves strong
performance despite its compact architecture and small number of
trainable parameters. This suggests that subjectless triples can be
effectively encoded in the semantic embedding space and aggregated with
predicate-aware weights to construct discriminative entity
representations for classification.

Second, the comparison between LLM and DRSD shows that pseudo-answer
distillation is effective. The distilled SLM outperforms the prompted
LLM and implicitly learns task-specific alignment rules embedded in the
data. We note that DRSD inherits a prior distribution biased toward
\textsc{same} due to the imbalanced training split (76.8\% \textsc{same})
and thus prioritizes recall, whereas PIE exhibits more balanced
performance. Finally, routing DRSD's uncertain predictions to human
review raises F1 from 0.886 to 0.909 (Table~\ref{tab:main_result}); we
analyze why this is possible in Section~\ref{sec:analysis}.

\subsection{Analysis}
\label{sec:analysis}
\paragraph{Predicate Budget $k$}
Table~\ref{tab:pred_k_ablation} reports prompt-only teacher LLM classification performance as the predicate budget $k$ in PIE's importance-based compression varies (Section~\ref{sec:PIW}). Performance changes
only marginally as $k$ grows from 10 to 20, while prompt length increases with $k$. This
indicates that the top-ranked predicates already carry most of the
alignment-relevant evidence, which serves as the basis for selecting $k{=}10$ to minimize the input volume while maintaining accuracy.

\begin{figure}[t]
  \centering
  \includegraphics[width=\columnwidth]{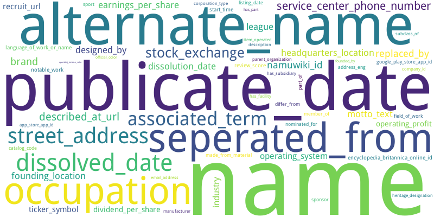}
  \caption{Predicate-importance weights learned by PIE for predicates in the specific domain (Company).}
  \label{fig:company_predicate_importance}
\end{figure}

\paragraph{Predicate-importance Weights}
Figure~\ref{fig:company_predicate_importance} visualizes the predicate-importance
weights that PIE learns in the company domain. The highly weighted
predicates correspond to attributes and relations describing
organizational identity or business context, indicating that PIE
learns interpretable predicate preferences consistent with human
judgment, beyond merely improving classification.

\begin{table}[t]
\centering
\small
\begin{tabular*}{\columnwidth}{@{\extracolsep{\fill}}ccccc@{}}
\toprule
\textbf{$k$} & \textbf{Acc.} & \textbf{Prec.} & \textbf{Rec.} & \textbf{F1} \\
\midrule
5  & 0.7956 & 0.7340 & \textbf{0.9272} & 0.8194 \\
10 & 0.7976 & 0.7433 & 0.9092 & 0.8179 \\
15 & 0.8044 & 0.7532 & 0.9056 & 0.8224 \\
20 & \textbf{0.8094} & \textbf{0.7574} & 0.9104 & \textbf{0.8269} \\
\bottomrule
\end{tabular*}
\caption{LLM classification results across top $k$ predicate budgets. Best scores are shown in bold.}
\label{tab:pred_k_ablation}
\end{table}

\paragraph{LGD vs DRSD}
Figure~\ref{fig:fn_fp} compares where LGD and DRSD place their errors
across score-bins (false negatives at top, false positives at bottom).
In LGD, the score is tied to the decision. \textsc{different}
predictions receive low scores and \textsc{same} predictions receive
high scores, so both false negatives and false positives for this model fall within the expected side of the threshold. As a result, LGD produces almost no
score--decision discrepancy, and no signal indicates which predictions
are unreliable. DRSD instead decouples the score from the decision, which can assign high scores to 24 pairs predicted as
\textsc{different} and low scores to 115 pairs predicted as \textsc{same}.

These discrepancies concentrate on the model's errors, and form a concrete criterion for human-in-the-loop
review. Only these conflicting cases (13.3\% of the evaluation set) need
to be routed to a human annotator. The same review applied to LGD
yields almost no change, since it produces no discrepancies to act on. In
an industrial KG-integration environment where false-positive merges result in significant costs and review capacity is limited, this allows an operator to automatically accept confident predictions and focus their review efforts solely on the small set of genuinely uncertain ones that a label-coupled score cannot support.
\pgfplotsset{
    mycustomstyle/.style={
        ybar,
        width=\columnwidth,
        height=4.0cm,
        ylabel={Count},
        ymin=0,
        xtick={0.0,0.1,0.2,0.3,0.4,0.5,0.6,0.7,0.8,0.9},
        bar width=3pt,
        tick label style={font=\scriptsize},
        label style={font=\small},
        ylabel style={overlay},
        ymajorgrids=true,
        grid style={draw=gray!20},
        title style={font=\bfseries\small, yshift=-1ex},
        legend style={font=\scriptsize, nodes={scale=0.8, transform shape}},
        axis background/.style={fill=white},
    }
}

\begin{figure}[htbp]
    \centering
    \begin{subfigure}[b]{\columnwidth}
        \centering
        \begin{tikzpicture}
            \begin{axis}[mycustomstyle, title={False Negative (FN)}, legend pos=north east, legend image code/.code={
        \draw[#1, draw=black] (0cm,-0.1cm) rectangle (0.1cm,0.1cm);
    }]
            \addplot[fill=blue!60, draw=black] coordinates {(0.0,1) (0.1,18) (0.2,30) (0.3,28) (0.4,13) (0.5,0) (0.6,1) (0.7,0) (0.8,0) (0.9,0)};
            \addlegendentry{LGD}
            \addplot[fill=red!60, draw=black] coordinates {(0.0,10) (0.1,13) (0.2,15) (0.3,6) (0.4,0) (0.5,0) (0.6,0) (0.7,4) (0.8,19) (0.9,1)};
            \addlegendentry{DRSD}
            \end{axis}
        \end{tikzpicture}
    \end{subfigure}

    \vspace{1mm}

    \begin{subfigure}[b]{\columnwidth}
        \centering
        \begin{tikzpicture}
            \begin{axis}[mycustomstyle, title={False Positive (FP)}]
            \addplot[fill=blue!60, draw=black] coordinates {(0.0,0) (0.1,0) (0.2,0) (0.3,0) (0.4,0) (0.5,0) (0.6,13) (0.7,289) (0.8,232) (0.9,3)};
            \addplot[fill=red!60, draw=black] coordinates {(0.0,11) (0.1,38) (0.2,43) (0.3,14) (0.4,9) (0.5,0) (0.6,9) (0.7,62) (0.8,188) (0.9,187)};
            \end{axis}
        \end{tikzpicture}
    \end{subfigure}
    \caption{Score-bin distribution of false negatives (top) and false
    positives (bottom). DRSD shows score--decision discrepancies (high-score
    FN, low-score FP) that can route to human review, whereas LGD produces
    almost none.}
    \label{fig:fn_fp}
\end{figure}
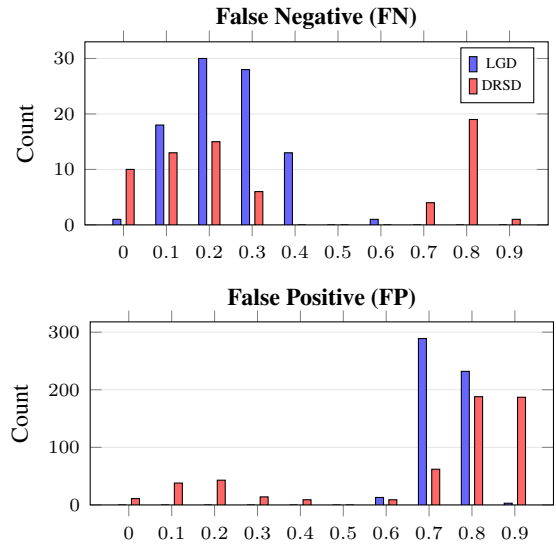

\section{Conclusion}

We presented the entity-alignment framework for heterogeneous KG
integration that combines Predicate Importance Estimation (PIE) and
Decoupled Rationale-Score Distillation (DRSD). PIE learns
predicate-aware entity representations from subjectless 1-hop triples,
while DRSD distills LLM reasoning into a smaller model by decoupling
label-consistent rationales from confidence-score estimation.
Crucially, because DRSD's score is decoupled from its decision,
discrepancies between decision and score detect likely errors and provide a criterion
for human review that a label-coupled score cannot offer. Correcting
only these detected error cases further improves F1. These results suggest that efficient and operationally practical EA is feasible by selecting
task-relevant graph evidence with a predicate-aware method and
distilling LLM reasoning into a compact model. The decoupling principle
is not limited to EA and it can also be useful in other human-in-the-loop
distillation settings.

\section*{Limitations}
Our study has several limitations.
First, although PIE is designed to learn predicate-importance weights,
these weights can become biased when some classes or predicates dominate
the training data. Ideally, predicate importance should be estimated in a
class-aware manner, but predicate distributions also vary across
collections. Future work should explore multiple importance weights,
ensemble-based weighting, or other debiasing strategies that produce
predicate weights generalizing across both classes and collections.

Second, we do not provide a component-level ablation of PIE, such as
isolating the contribution of subjectless encoding or comparing learnable
importance weights against uniform pooling. A finer-grained analysis of
these design choices is left to future work.

Third, our human-review results assume correct annotator judgment and
therefore represent an upper bound on the achievable gain. In practice,
the realized improvement depends on annotator accuracy and the
available review budget, and the 13.3\% of pairs flagged for review may
be costly to inspect at scale. Estimating the gain under realistic
annotator behavior is left to future work.
\section*{Ethical Considerations}

\paragraph{Data.} Our experiments use proprietary, in-house knowledge
graphs constructed from internal databases. The data contains entity
records such as organizations, people, and creative works but no private
personal information beyond what is publicly associated with these
entities. Because the data is proprietary, we do not release it; we
instead report dataset statistics and full
implementation details (Appendix~\ref{sec:data_statistics}) to support
reproducibility of the method.

\paragraph{Human review.} The ``+ Human Review'' row in
Table~\ref{tab:main_result} uses the manually verified evaluation labels
as the outcome of reviewing the flagged score--decision conflicts.
Because these labels are human-verified, the result can be interpreted
as a simulation of correct human review on the flagged subset. We report
it separately from fully automatic methods because deployment gains
would still depend on annotator accuracy, review capacity, and
quality-control procedures.

\paragraph{Intended use.} The proposed method targets knowledge-graph
integration in industrial settings. As with any automated alignment
system, incorrect merges could propagate errors into downstream
applications. The confidence-based routing we propose is designed
precisely to mitigate this risk by escalating uncertain cases to human
review.



\bibliography{custom}

\appendix

\section{Data Statistics}
\label{sec:data_statistics}

\paragraph{Knowledge Graphs} Our entity pairs are drawn from in-house heterogeneous knowledge graphs spanning multiple domains and sources. The graphs encompass entities from diverse domains such as entertainment (films, music, and broadcasts), companies, people, and cultural properties, connected through 2,904 distinct relation types. Because the graphs are heterogeneous, entities vary in their types and in the relations connecting them, so the set of 1-hop triples available for an entity differs substantially in both quantity and schema across the dataset. For each entity in a pair, we extract its 1-hop triples as the structural context provided to the model.

\paragraph{Statistics} Table~\ref{tab:dataset_statistics} reports the dataset scale, label distribution, and 1-hop triple statistics for the training and evaluation splits. The labeled training split is imbalanced toward \textsc{same} pairs, which account for 76.8\% of training examples, whereas the evaluation split is intentionally balanced with a 50/50 label ratio. For the triple statistics, Triples/entity (mean) indicates the average number of 1-hop triples available per entity, while Triples/entity (max) indicates the largest number of 1-hop triples observed in each split.

\begin{table}[!h]
\centering
\small
\begin{tabular*}{\columnwidth}{@{\extracolsep{\fill}}lrr@{}}
\hline
\textbf{Statistic} & \textbf{Train} & \textbf{Evaluation} \\
\hline
Entity pairs & 85,858 & 5,000 \\
\textsc{same} & 65,984 (76.8\%) & 2,500 (50.0\%) \\
\textsc{different} & 19,874 (23.2\%) & 2,500 (50.0\%) \\
Unique entities & 136,626 & 9,499 \\
Triples/entity (mean) & 26.17 & 25.0 \\
Triples/entity (max) & 4,349 & 1,811 \\
\hline
\end{tabular*}
\caption{Dataset statistics for the labeled training and evaluation splits. Triples/entity reports the mean and maximum number of 1-hop triples per entity.}
\label{tab:dataset_statistics}
\end{table}

\section{Implementation Details}
\label{sec:implementation_details}

\subsection{Decoupled Rationale-Score Distillation}
We fine-tune an hcx-srch1-8B model as the student SLM using the SFT objective described in the main method, with DeepSpeed ZeRO Stage 2, \texttt{bfloat16} precision, and gradient checkpointing. We train for 1 epoch with AdamW (learning rate $2\times10^{-5}$, 100 warmup steps, no weight decay), an effective batch size of 8, and a maximum sequence length of 4096 tokens. We apply completion-only loss masking, discard over-length samples, and use greedy decoding with up to 256 new tokens at inference time. During inference, the decision field is used for match classification, and the score field is retained for threshold-based confidence analysis.

\subsection{Predicate Importance Estimation (PIE)}
All predicate--object texts are encoded offline with the \texttt{BAAI/bge-m3} encoder. For each candidate pair, the resulting entity embeddings are concatenated and passed to a two-layer MLP classifier. The pretrained encoder is frozen, and only the predicate-importance weights and MLP classifier layers are trained with binary cross-entropy using AdamW (learning rate $1\times10^{-3}$, 100 warmup steps, no weight decay) for up to 20 epochs; validation accuracy selects the best checkpoint.

\subsection{Pseudo-Answer Construction}
\label{sec:PA construction}
Each entity's 1-hop neighborhood is serialized as JSON-style evidence, excluding collection-type objects and retaining at most 10 objects per predicate to control prompt length. For each labeled candidate pair, the teacher LLM (\texttt{openai/gpt-oss-120b}) is queried with both the gold-label prompt and the score prompt. The final SFT target follows the main decoupled rationale-score construction: the decision and justification are taken from the gold-label output, while the score is taken from the score prompt output.
\clearpage
\onecolumn
\section{Prompt \& Output Examples}
\label{sec:prompt_examples}

    \small
    
    \begin{tcolorbox}[
        colback=gray!10!white,
        colframe=black!75!black,
        title=\textbf{(A) Gold-Label Prompt (GOLD\_TEMPLATE)},
        width=\textwidth,
        boxrule=1pt
    ]
        You will receive two JSON descriptions, and each description is generated from an entity's subgraph (attributes and relations).
        
        \vspace{2mm}
        
        You will ALSO receive the manually verified gold (ground-truth) decision label.
        
        \vspace{2mm}
        
        Your task is to generate: \\
        1) A well-calibrated similarity score in [0, 1] \\
        2) A high-quality justification that is CONSISTENT with the given decision and the score that you generate
        
        \vspace{1mm}
        
        This output will be used for knowledge distillation.
        
        \vspace{2mm}
        
        Your task has TWO phases:
        
        \vspace{2mm}
        
        \textbf{PHASE 1 (internal reasoning, not fully exposed):} \\
        - Use the entity instance type that the model infers to determine which features are CORE vs. AUXILIARY. \\
        - Compare the two entities with type-appropriate core attributes and relations. \\
        - Identify matches that provide support, explainable discrepancies, conflicts, and information that is missing. \\
        - Choose a similarity score that: \\
        ~~- Is consistent with the evidence strength \\
        ~~- Respects the decision threshold: \\
        ~~~~- If decision = "SAME", score MUST be $\geq$ 0.60 \\
        ~~~~- If decision = "DIFFERENT", score MUST be $<$ 0.60 \\
        - Construct reasoning that supports BOTH the decision and the chosen score.
        
        \vspace{2mm}
        
        \textbf{IMPORTANT:} \\
        - You MUST NOT contradict the gold decision that the prompt provides. \\
        - The score that you generate MUST be consistent with the decision. \\
        - Treat the gold decision as correct, even if the evidence is weak or ambiguous.
        
        \vspace{2mm}
        
        \textbf{PHASE 2 (final output only):} \\
        - Output ONLY the simplified JSON that the template below defines. \\
        - Do NOT include internal reasoning or mention the gold decision explicitly.
        
        \vspace{2mm}
        
        \textbf{[Entity A Description]} \\
        \{A\_TEXT\}
        
        \vspace{1mm}
        
        \textbf{[Entity B Description]} \\
        \{B\_TEXT\}
        
        \vspace{1mm}
        
        \textbf{[Gold Label]} \\
        Decision: \{GOLD\_DECISION\}
        
        \vspace{2mm}
        
        \textbf{FINAL OUTPUT FORMAT (STRICT):} \\
        Return ONLY valid JSON in the following format, with no additional text.
        
        \vspace{1mm}
        
        \texttt{\{} \\
        \texttt{~~"justification": [} \\
        \texttt{~~~~"Brief reason 1 (most important type-specific evidence or explanation)",} \\
        \texttt{~~~~"Brief reason 2",} \\
        \texttt{~~~~"Brief reason 3"} \\
        \texttt{~~],} \\
        \texttt{~~"decision": "\{GOLD\_DECISION\}",} \\
        \texttt{~~"score": 0.0} \\
        \texttt{\}}
        
        \vspace{2mm}
        
        \textbf{Constraints:} \\
        - The decision MUST exactly match the gold decision that the prompt provides. \\
        - The score MUST be a float in [0, 1] with at most two decimal places. \\
        - Score constraints: \\
        ~~- If decision = "SAME" $\rightarrow$ score $\geq$ 0.60 \\
        ~~- If decision = "DIFFERENT" $\rightarrow$ score $<$ 0.60 \\
        - Justifications must be: \\
        ~~- Type-appropriate \\
        ~~- Consistent with both the decision and the score magnitude \\
        - Do NOT add or remove fields.
    \end{tcolorbox}
    
    \captionof{figure}{Gold-label prompt used to generate a label-consistent decision and justification for distillation.}
    \label{fig:triple_prompt}

\begin{figure*}[t!]
    \centering
    \small
    
    \begin{tcolorbox}[
        colback=gray!10!white,
        colframe=black!75!black,
        title=\textbf{(B) Score Prompt (SCORE\_TEMPLATE)},
        width=\textwidth,
        boxrule=1pt
    ]
        You will receive two JSON descriptions, and each description is generated from an entity's subgraph (attributes and relations).
        
        \vspace{2mm}
        
        Your task has TWO phases:
        
        \vspace{2mm}
        
        \textbf{PHASE 1 (internal reasoning, not fully exposed):} \\
        - Compare the two entities with the most type-appropriate core attributes and relations. \\
        - Identify high-confidence matches, matches that provide support, conflicts (high/medium/low), and information that is missing or uncertain. \\
        - conflicts should be weighted by their importance to the entity type. \\
        - Apply the scoring principles below to compute a similarity score between 0 and 1.
        
        \vspace{2mm}
        
        \textbf{PHASE 2 (final output only):} \\
        - Output ONLY the simplified JSON that the template below defines. \\
        - Do NOT include the full internal analysis in the final answer.
        
        \vspace{2mm}
        
        - Unique identifiers $>$ entity type $>$ type-specific core attributes $>$ key relations $>$ auxiliary attributes. \\
        - Type mismatch or unique ID conflicts should strongly reduce the score. \\
        - Missing information increases uncertainty but is not a conflict.
        
        \vspace{2mm}
        
        \textbf{[Entity A Description]} \\
        \{A\_TEXT\}
        
        \vspace{1mm}
        
        \textbf{[Entity B Description]} \\
        \{B\_TEXT\}
        
        \vspace{2mm}
        
        \textbf{FINAL OUTPUT FORMAT (STRICT):} \\
        Return ONLY valid JSON in the following format, with no additional text.
        
        \vspace{1mm}
        
        \texttt{\{} \\
        \texttt{~~"justification": [} \\
        \texttt{~~~~"Brief reason 1 (most important type-specific evidence or conflict)",} \\
        \texttt{~~~~"Brief reason 2",} \\
        \texttt{~~~~"Brief reason 3"} \\
        \texttt{~~],} \\
        \texttt{~~"score": 0.0} \\
        \texttt{\}}
        
        \vspace{2mm}
        
        \textbf{Constraints:} \\
        - The score must be a float in [0, 1] with at most two decimal places. \\
        - Justification items must reflect features that are appropriate for the entity type that the model infers.
    \end{tcolorbox}
    
    \caption{Score prompt used to estimate an evidence-based similarity score for an entity pair.}
    \label{fig:score_prompt}
\end{figure*}

\begin{figure}[t!]
    \centering
    \small
    
    \begin{tcolorbox}[
        colback=gray!10!white,
        colframe=black!75!black,
        title=\textbf{(C) SFT Prompt (SFT\_TEMPLATE)},
        width=\columnwidth,
        boxrule=1pt
    ]
        You will receive two JSON descriptions, and each description is generated from an entity's subgraph (attributes and relations).
        
        \vspace{2mm}
        
        Determine whether the two entities refer to the SAME entity or DIFFERENT entities.
        
        \vspace{2mm}
        
        \textbf{[Entity A Description]} \\
        \{A\_TEXT\}
        
        \vspace{1mm}
        
        \textbf{[Entity B Description]} \\
        \{B\_TEXT\}
        
        \vspace{2mm}
        
        \textbf{Output:} \\
        - decision: SAME or DIFFERENT \\
        - score: a similarity score in [0, 1]\\
        - three brief justification points
        
        \vspace{2mm}
        
        Return only valid JSON in the specified format.
    \end{tcolorbox}
    
    \caption{SFT prompt used as the student-model input at training and inference time.}
    \label{fig:sft_prompt}
\end{figure}
\begin{figure}[t!]
    \centering
    \small
    
    \begin{tcolorbox}[
        colback=gray!10!white,
        colframe=black!75!black,
        title=\textbf{(A) Gold-Label Output (JSON)},
        width=\columnwidth,
        boxrule=1pt
    ]
        \texttt{\{} \\
        \texttt{~~"justification": [} \\
        \texttt{~~~~"Both entities share the title \textbackslash"Memories\textbackslash" but have different release years (2023 vs 2019).",} \\
        \texttt{~~~~"The listed composers and performers differ completely (J.Fla and multiple Western composers vs Maroon~5 and Korean-named composers).",} \\
        \texttt{~~~~"Additional metadata such as record label, language, and related works are unrelated. This indicates that they are distinct musical works."} \\
        \texttt{~~],} \\
        \texttt{~~"decision": "DIFFERENT",} \\
        \texttt{~~"score": 0.22} \\
        \texttt{\}}
    \end{tcolorbox}
    
    \caption{Example gold-label prompt output containing a decision, score, and justification.}
    \label{fig:example_output}
\end{figure}

\begin{figure}[t!]
    \centering
    \small
    
    \begin{tcolorbox}[
        colback=gray!10!white,
        colframe=black!75!black,
        title=\textbf{(B) Score Prompt Output (JSON)},
        width=\columnwidth,
        boxrule=1pt
    ]
        \texttt{\{} \\
        \texttt{~~"justification": [} \\
        \texttt{~~~~"Both entities share the exact name \textbackslash"Memories\textbackslash" and overlapping core composers (Adam Levine and Jonathan/Jon Bellion). This indicates the same musical work.",} \\
        \texttt{~~~~"Entity A is a 2023 cover by J.Fla, whereas Entity B is the original 2019 recording by Maroon 5; the performer and release-date differences reflect different recordings, not a conflict on composition identity.",} \\
        \texttt{~~~~"Other attributes (record label, part\_of album, instance\_of) differ but are performance-specific and do not outweigh the strong composer and title match."} \\
        \texttt{~~],} \\
        \texttt{~~"score": 0.78} \\
        \texttt{\}}
    \end{tcolorbox}
    
    \caption{Example score prompt output containing an evidence-based score and justification.}
    \label{fig:example_output_same}
\end{figure}

\begin{figure}[t!]
    \centering
    \small
    
    \begin{tcolorbox}[
        colback=gray!10!white,
        colframe=black!75!black,
        title=\textbf{(C) SFT Target (JSON)},
        width=\columnwidth,
        boxrule=1pt
    ]
        \texttt{\{} \\
        \texttt{~~"justification": [} \\
        \texttt{~~~~"Both entities share the title \textbackslash"Memories\textbackslash" but have different release years (2023 vs 2019).",} \\
        \texttt{~~~~"The listed composers and performers differ completely (J.Fla and multiple Western composers vs Maroon~5 and Korean-named composers).",} \\
        \texttt{~~~~"Additional metadata such as record label, language, and related works are unrelated. This indicates that they are distinct musical works."} \\
        \texttt{~~],} \\
        \texttt{~~"decision": "DIFFERENT",} \\
        \texttt{~~"score": 0.78} \\
        \texttt{\}}
    \end{tcolorbox}
    
    \caption{Example SFT target combining the gold-label decision and justification with the score prompt score.}
    \label{fig:drsd_label}
\end{figure}

\end{document}